\documentclass[letterpaper]{article} % DO NOT CHANGE THIS
\usepackage{aaai2026}  % DO NOT CHANGE THIS
\usepackage{times}  % DO NOT CHANGE THIS
\usepackage{helvet}  % DO NOT CHANGE THIS
\usepackage{courier}  % DO NOT CHANGE THIS
\usepackage[hyphens]{url}  % DO NOT CHANGE THIS
\usepackage{graphicx} % DO NOT CHANGE THIS
\usepackage{natbib}  % DO NOT CHANGE THIS AND DO NOT ADD ANY OPTIONS TO IT
\usepackage{caption} % DO NOT CHANGE THIS AND DO NOT ADD ANY OPTIONS TO IT
\usepackage{algorithm}
\usepackage{algorithmic}

\usepackage{amsmath}
\usepackage{amsfonts}
\usepackage{pifont}
\usepackage{xcolor}
\usepackage{newfloat}
\usepackage{listings}
\DeclareCaptionStyle{ruled}{labelfont=normalfont,labelsep=colon,strut=off} % DO NOT CHANGE THIS
\floatstyle{ruled}
\newfloat{listing}{tb}{lst}{}
\floatname{listing}{Listing}
\title{Topology Matters: A Cautionary Case Study of Graph SSL on Neuro-Inspired Benchmarks}
\author {
    May Kristine Jonson Carlon\textsuperscript{\rm 1},
    Su Myat Noe\textsuperscript{\rm 2},
    Haojiong Wang\textsuperscript{\rm 1},
    Yasuo Kuniyoshi\textsuperscript{\rm 1}
}
\affiliations {
    \textsuperscript{\rm 1}RIKEN Center for Brain Science, Japan\\
    \textsuperscript{\rm 2}National Institute of Informatics, Japan\\
}

\begin{document}

\maketitle

\begin{abstract}
Understanding how local interactions give rise to global brain organization requires models that can represent information across multiple scales. We introduce a hierarchical self-supervised learning (SSL) framework that jointly learns node-, edge-, and graph-level embeddings, inspired by multimodal neuroimaging. We construct a controllable synthetic benchmark mimicking the topological properties of connectomes. Our four-stage evaluation protocol reveals a critical failure: the invariance-based SSL model is fundamentally misaligned with the benchmark's topological properties and is catastrophically outperformed by classical, topology-aware heuristics. Ablations confirm an objective mismatch: SSL objectives designed to be invariant to topological perturbations learn to ignore the very community structure that classical methods exploit. Our results expose a fundamental pitfall in applying generic graph SSL to connectome-like data. We present this framework as a cautionary case study, highlighting the need for new, topology-aware SSL objectives for neuro-AI research that explicitly reward the preservation of structure (e.g., modularity or motifs).
\end{abstract}

\section{Introduction}

Biological intelligence emerges from hierarchically organized neural systems, where local microcircuits process sensory features and long-range connections integrate information across distributed networks. Modern neuroimaging and connectomics offer unprecedented access to this multiscale structure, revealing that brain organization can be naturally modeled as a \emph{multiplex graph}: nodes correspond to cortical or subcortical regions, and edges capture structural or functional relationships across modalities such as morphometry, diffusion tractography, and resting-state functional magnetic resonance imaging (fMRI)~\cite{vanessen2013wuminn,parisot2017spectral}. These multimodal connectomes embody three key properties: (1) rich node-level features reflecting regional anatomy and function; (2) multi-channel edges encoding diverse connectivity modalities; and (3) higher-order organization---such as communities, hubs, and hemispheric symmetry---that supports distributed computation. Capturing all three scales simultaneously is essential for both understanding neural systems and developing biologically grounded machine learning architectures~\cite{ktena2018metric,liu2024objective}.

\paragraph{From Neuro to AI.}
Principles of cortical computation --- such as hierarchical predictive coding, sparse and energy-efficient representations, and structured connectivity --- motivate learning frameworks that balance invariance and selectivity across representational levels. Graph Neural Networks (GNNs) provide a natural substrate for exploring these ideas because their architecture natively handles graph-structured data, making them perfectly suited to model the multiscale organization inherent in the brain's connectome. Self-supervised learning (SSL) is a valuable paradigm for this domain as it enables learning rich representations from the vast amount of unlabeled connectomic data; however, most existing graph SSL approaches are limited because they focus only on node-level or graph-level objectives, neglecting the rich relational information carried by the edges. As a result, current graph SSL methods cannot fully capture how local interactions shape global organization, an ability critical for both neural modeling and robust AI systems.

\paragraph{From AI to Neuro.}
Conversely, SSL offers neuroscientists a scalable means to learn task-agnostic representations by uncovering the latent structure of brain graphs, potentially serving as a computational analog of unsupervised cortical learning and aiding the discovery of functional modules. However, current graph SSL methods face major limitations. Contrastive approaches like Deep Graph Infomax (DGI)~\cite{veličković2018deep} and GraphCL~\cite{you2020graph} typically restrict learning to a single representational scale. Non-contrastive methods such as VICReg~\cite{bardes2022vicreg} and Barlow Twins~\cite{zbontar2021barlow} fail to explicitly model hierarchical dependencies among nodes, edges, and graphs, thereby neglecting the pairwise interactions (edge embeddings) central to understanding functional and structural coupling. Moreover, evaluation in graph SSL often lacks statistical rigor~\cite{errica2022faircomparison,zhu2024pitfalls,gong2025identifying}, as inconsistent benchmarks and ad hoc tuning hinder the assessment of architectural components. To address this methodological gap, we first develop a controllable, neuro-inspired synthetic benchmark for standardized evaluation and then use it to conduct a failure analysis of a representative hierarchical SSL framework. Our contributions are:

\subsection{Contributions}
\paragraph{Hierarchical SSL with explicit edge modeling.} 
We propose a dedicated edge projection head that combines endpoint and multimodal attribute information, producing explicit, queryable edge embeddings. SimSiam-style predictors~\cite{chen2021exploring} enforce cross-view alignment at each representational scale, maintaining architectural simplicity while preventing collapse.

\paragraph{Failure analysis of invariance-based SSL.}
Through systematic ablations, we demonstrate a fundamental objective mismatch. We show that modern invariance-based SSL objectives, which are designed to discard topological details, are outperformed by simple heuristics that exploit the community structure inherent in our neuro-inspired benchmark.

\paragraph{Rigorous Evaluation.}
Our four-stage protocol includes principled hyperparameter optimization, multi-task probing with statistical significance tests, transfer evaluation on unseen graphs, and component-level ablations, offering a transparent methodology for future multi-scale SSL research.

\paragraph{Neuro-inspired Synthetic Benchmark.}
We introduce a controllable multiplex graph generator that mimics multimodal neuroimaging properties, allowing controlled evaluation and hypothesis testing. This benchmark provides a bridge between biologically inspired modeling and scalable graph learning.

\subsection{Related Work}

\paragraph{Graph Self-Supervised Learning.}
Contrastive methods like DGI~\cite{veličković2018deep} and GraphCL~\cite{you2020graph} maximize agreement between augmented views but focus on single scales and require careful negative sampling. Non-contrastive approaches like VICReg~\cite{bardes2022vicreg} and BGRL~\cite{thakoor2023large} avoid collapse without negatives but do not explicitly model edges.

\paragraph{Edge and Hierarchical Learning.}
Few SSL works address edge embeddings. GraphMAE~\cite{hou2022graphmae} masks edges for reconstruction but lacks explicit edge representations. GMT~\cite{baek2021accurate} learns hierarchical features but requires supervision. Graph pre-training work~\cite{hu2020strategies,qiu2020gcc,li2021pairwise,liu2022pretraining} explores multi-scale strategies but is categorized into microscopic, mesoscopic, and macroscopic paradigms~\cite{zhao2025survey}. Our work uniquely integrates explicit edge embeddings into a fully self-supervised hierarchical framework with rigorous multi-scale evaluation.

\paragraph{Pre-training of Graph Neural Networks.}
Early work on graph pre-training demonstrated that general-purpose GNN encoders could benefit downstream tasks when trained on unlabeled graphs through structural and contextual objectives~\cite{hu2020strategies}. Subsequent approaches introduced contrastive frameworks that discriminate between subgraph pairs across multiple networks~\cite{qiu2020gcc} or between augmented halves of the same graph~\cite{li2021pairwise}, enabling new transferable representations even across heterogeneous domains. More recent studies explore multi-view or multi-scale pre-training strategies that jointly capture local and global semantics, as in molecular~\cite{liu2022pretraining} and heterogeneous graph domains. Comprehensive surveys~\cite{zhao2025survey} categorize these efforts into microscopic (node-level), mesoscopic (edge- or subgraph-level), and macroscopic (graph-level) paradigms. Our approach falls within this line of work but extends it to a hierarchical setting.

\section{Method}

\subsection{Problem Setup}

We consider a multiplex graph $G = (V, E, X, A)$ where $V$ is a set of $N$ nodes with feature matrix $X \in \mathbb{R}^{N \times F}$, $E \subseteq V \times V$ is the union edge set, and $A \in \mathbb{R}^{M \times C}$ represents $M$ edges with $C$ channels (e.g., functional connectivity [SC] weight, functional connectivity [FC] correlation). Our goal is to learn an encoder $f_\theta$ producing:
\begin{align}
    \{z_v\}_{v \in V} &\in \mathbb{R}^{N \times D_n} \quad \text{(nodes)} \\
    \{z_{uv}\}_{(u,v) \in E} &\in \mathbb{R}^{M \times D_e} \quad \text{(edges)} \\
    z_G &\in \mathbb{R}^{D_g} \quad \text{(graph)}
\end{align}
\noindent
These representations should be invariant to semantics-preserving augmentations while capturing task-relevant information at each scale. All modalities are represented as multi-channel edge attributes, preserving modality-specific signals within shared message passing.

\subsection{Architecture}

\paragraph{Shared GNN backbone.}
We use GraphSAGE-style message passing~\cite{hamilton2017inductive} with $L$ layers:
\begin{equation}
h_v^{(\ell)} = \sigma\left(W_s^{(\ell)} h_v^{(\ell-1)} + W_n^{(\ell)} \text{MEAN}(\{h_u^{(\ell-1)} : u \in \mathcal{N}(v)\})\right)
\end{equation}
where $h_v^{(0)} = x_v$. This produces hidden states $\{h_v^{(L)}\}$.
Note that this message passing uses node features and topology but does not explicitly incorporate the multi-channel edge attributes during aggregation.

\paragraph{Multi-level projection heads.}
For nodes and graphs, we use 2-layer MLPs with ReLU~\cite{nair2010rectified} and LayerNorm~\cite{ba2016layernormalization}:
\begin{align}
z_v &= \mathrm{MLP}_{\text{node}}(h_v^{(L)}),
\end{align}
and an attention-style readout with learnable scalar query $q$ for the graph embedding:
\begin{align}
s_v &= q^\top h_v^{(L)}, \\
\text{(train)}\quad w_v^{\text{train}} &= \mathrm{softmax}\!\left(\frac{s_v}{\tau}\right),\quad \tau{=}5, \\
\text{(eval)}\quad w_v^{\text{eval}} &= \frac{\sigma(s_v)}{\sum_{u}\sigma(s_u)}, \\
z_G &= \mathrm{MLP}_{\text{graph}}\!\left(\sum_{v} w_v\, h_v^{(L)}\right).
\end{align}

For edges, we concatenate endpoint hidden states with a learned edge-attribute embedding:
\begin{equation}
z_{uv} = \mathrm{MLP}_{\text{edge}}\!\big([\,h_u^{(L)};\,h_v^{(L)};\,\phi(a_{uv})\,]\big).
\end{equation}
Here, $\mathbf{\phi(a_{uv})}$ is a learned embedding of the multi-channel edge attributes, enabling the model to utilize relational information at the final projection stage. Unlike latent messages implicitly formed within GNN layers, our $z_{uv}$ are explicitly learned edge representations accessible to downstream edge-level tasks.

\paragraph{SimSiam predictors.}
Following SimSiam~\cite{chen2021exploring}, we add predictor MLPs at each scale:
\begin{equation}
p_{\text{scale}} = \text{Predictor}(z_{\text{scale}})
\end{equation}
\noindent
These asymmetric predictors enable one-sided gradient flow, preventing collapse without negative pairs.

\subsection{Self-Supervised Learning Objective}

We create two augmented views $G^A, G^B$ and minimize:
\begin{equation}
    \mathcal{L} = \sum_{\text{scale} \in \{n,e,g\}} \lambda_{\text{scale}} \mathcal{L}_{\text{scale}} + \mathcal{L}_{\text{reg}}
\end{equation}

We set $\lambda_{\mathrm{PN}}=\lambda_{\mathrm{PE}}=\lambda_{\mathrm{PG}}=1.0$
for the SimSiam predictor terms on nodes/edges/graphs. The edge distribution term uses $\lambda_E \in \{0.5,\,1.0,\,2.5\}$ during tuning, and the tuned winner is used for all downstream results. Variance and covariance regularizers follow VICReg-style magnitudes: $\alpha=0.1$ for each variance term and $\beta=0.15$ for each covariance term (i.e., the objective contains $0.1\,[\mathrm{Var}(z_n^A)+\mathrm{Var}(z_n^B)+\mathrm{Var}(z_e^A)+\mathrm{Var}(z_e^B)]$ and $0.15\,[\mathrm{Cov}(z_n^A)+\mathrm{Cov}(z_n^B)+\mathrm{Cov}(z_e^A)+\mathrm{Cov}(z_e^B)]$). We also add a small invariance MSE on nodes ($\mathcal{L}_{\mathrm{inv},n}$) with unit weight.

\paragraph{View augmentation.} 
\begin{description}
    \item[Node feature masking.] Drop features with probability $p=0.02$. 
    \item[DropEdge.] Randomly retain 85\% of edges per view. DropEdge (keep 0.85) sparsifies each view; the base union graph is connected, but individual views need not remain connected. It also acts as a mild regularizer, preserving community structure while encouraging robustness.
\end{description}

\paragraph{Per-scale predictor losses.} 
For nodes and graphs, we use asymmetric SimSiam loss:
\begin{equation}
    \mathcal{L}_{\text{node}} = \frac{1}{2}\left[\ell(p_n^A, z_n^B) + \ell(p_n^B, z_n^A)\right]
\end{equation}
where $\ell(p, z) = -\langle p, \text{sg}(z) \rangle$ with stop-gradient $\text{sg}(\cdot)$ and $\ell_2$ normalization. We apply analogous predictor losses at all three scales. Additionally, we include a small MSE invariance term on nodes for stability, while edges are aligned distributionally via MMD.

\paragraph{Edge distribution matching.} 
Since augmentation changes edge sets, we align distributions via Maximum Mean Discrepancy (MMD) with RBF kernels~\cite{gretton2012kernel}:
\begin{equation}
    \mathcal{L}_{\text{edge-dist}} = \text{MMD}_{\text{RBF}}(Z_E^A, Z_E^B)
\end{equation}
where
\begin{align}
\mathrm{MMD}^2(Z_E^A,Z_E^B)
&=\frac{1}{|\Sigma|}\sum_{\sigma\in\Sigma}\!\Big[\,
  \mathbb{E}\,k_\sigma(z^A,z^{A'})  \notag\\
&\quad+\mathbb{E}\,k_\sigma(z^B,z^{B'})  \notag\\
&\quad-2\,\mathbb{E}\,k_\sigma(z^A,z^B)\Big].
\end{align}
\noindent
Here $k_\sigma(\mathbf{z},\mathbf{z}')=\exp\!\big(-\|\mathbf{z}-\mathbf{z}'\|^2/(2\sigma^2)\big)$,
$\Sigma=\{0.5,1.0,2.0\}$, and expectations are estimated with up to 4096 edges per view. MMD offers gradient stability under varying edge counts, unlike pairwise contrastive losses that require fixed correspondences.

\paragraph{Variance/covariance regularization.} 
To prevent collapse, we add VICReg-style~\cite{bardes2022vicreg} regularization:
\begin{align}
    \mathcal{L}_{\text{var}} &= \sum_d \text{ReLU}(\gamma - \sqrt{\text{Var}(Z_d) + \epsilon}) \\
    \mathcal{L}_{\text{cov}} &= \sum_{i \neq j} C_{ij}^2
\end{align}
applied to node and edge embeddings with $\gamma=0.2$, $\epsilon=10^{-6}$.

\paragraph{Training.} 

We use Adam (lr $=10^{-3}$, weight decay $=10^{-5}$), gradient clipping at $1.0$, and at most $400$ epochs with early stopping (patience $=6$, min\_delta $=10^{-3}$). We set \texttt{steps\_per\_epoch}$=1$, feature masking probability $p=0.02$, \texttt{DropEdge} keep ratio $=0.85$, and enable edge-embedding normalization during training (\texttt{normalize\_edges} = True).

\section{Experimental Setup}

\subsection{Synthetic Benchmark}

We generate 500 brain-inspired multiplex graphs with controllable properties to isolate representation effects free from data leakage or site confounds. To verify that our benchmark emulates key topological properties of real connectomes, we computed the average clustering coefficient ($C$) of the generated graphs and average shortest path length ($L$). Our benchmark graphs exhibit a high $C$ ($0.259 \pm 0.032$) and a low $L$ ($1.87 \pm 0.025$), relative to random graph equivalents. These are the classic hallmarks of the ``small-world'' topology widely observed in real structural and functional brain networks~\cite{watts1998collective,bullmore2009complex}. This confirms our benchmark serves as a valid, controlled ``model organism'' for testing SSL objectives on brain-like graph structures. Future work will extend to real MRI connectomes.

\paragraph{Topology.}
Each graph has $N \sim \text{Unif}(700, 900)$ nodes organized into $K \sim \text{Unif}(6, 10)$ communities via latent space models. Nodes are assigned to 3 simulated acquisition sites for batch effects.

\paragraph{Latent structure.}
Each community $k$ has two latent spaces: $Z_A \in \mathbb{R}^{3}$ (morphometric features) and $Z_B \in \mathbb{R}^{3}$ (connectivity/microstructure). Nodes sample from community-specific Gaussians.

\paragraph{Node features.}
($F=6$): Volume, thickness (log-normal from $Z_A$), FA, MD (sigmoid from $Z_B$), plus two auxiliary features. Site-specific offsets model batch effects.

\paragraph{Edges.}
SC edges form via: $P(\text{edge}) = \sigma(\beta_{\text{comm}} \mathbb{1}_{\text{same}} + \beta_{\text{sim}} \text{cos}(z_u^B, z_v^B) + b)$ with log-normal weights. We simulate 25\% missing SC. FC edges come from bandpass-filtered ($0.10$-$0.20$ Hz) AR(1) time series with community drivers, retaining top-30 correlations per node. Final multiplex has $M \approx 6000$-$8000$ edges with 2-channel attributes [SC weight, FC correlation].

\paragraph{Labels.}
One graph is selected for single-graph downstream probing. We design composite tasks requiring multimodal integration:
\begin{itemize}
    \item \emph{Node classification} (3 classes): Combines PageRank and feature score, discretized via quantiles.
    \item \emph{Link prediction}: 15\% held-out edges (positive) + equal negatives.
    \item \emph{Subgraph regression}: 200 random subgraphs scored by density and conductance.
\end{itemize}

All tasks use stratified 70/15/15 train/val/test splits.

\subsection{Four-Stage Evaluation Protocol}

\paragraph{Stage 1: Hyperparameter search.}
Grid search over architecture ($\text{hidden} \in \{64, 128\}$, $\text{depth} \in \{2, 3\}$, $\text{emb\_dim} \in \{32, 64\}$) and loss weights ($\lambda_E \in \{0.5, 1.0, 2.5\}$) yields 24 configs. Each trains for 400 epochs (early stopping). We compute composite validation scores:
\begin{equation}
    S_{\text{comp}} = \sum_{\text{task}} \frac{\text{score}_{\text{task}} - \min}{\max - \min}
\end{equation}

We tune on a single reference graph (the first synthesized instance): pre-train the encoder on the training portion of that graph only, compute validation-only probes (node/edge/subgraph) on its held-out validation split, and select the configuration that maximizes the composite score $S_{\text{comp}}$. The chosen hyperparameters and weights are then frozen and used everywhere else (single-graph test probes, transfer, ablations). Test sets are never consulted during model or hyperparameter selection.

\paragraph{Stage 2: Single-graph probes.}
Our method (``Ours'') uses the output of the pre-trained encoder to solve the downstream tasks. The encoder's parameters are \emph{frozen} during probing, and simple machine learning models (probes) are trained on top of the resulting embeddings. The methodology differs slightly for each task:

\begin{description}
    \item[Node Classification] 1-hidden-layer MLP (128 units) on frozen $z_v$.
    \item[Graph Regression] Ridge on mean-pooled edge embeddings within each subgraph. We report test scores and paired bootstrap significance versus the best baseline ($n{=}2000$).
    \item[Link Prediction] train/val edges (disjoint from test) are scored by a logistic-regression probe on \emph{edge embeddings produced on-the-fly} with the frozen edge head: $z_{uv}=\mathrm{MLP}_{\mathrm{edge}}([h_u;h_v;0])$ where $h=\text{backbone}(X,E)$ is computed once.
\end{description}

We compare ``Ours'' against three classes of baselines:

\begin{description}
\item[Classical.] These methods ignore graph topology. For link prediction, we use \textbf{Cosine Similarity}, scoring links $(u, v)$ by the cosine of the angle between their raw feature vectors $x_u$ and $x_v$. For subgraph regression, we use \textbf{Ridge(pool)}, where all node features in a subgraph are mean-pooled into a single vector for a standard Ridge regression. For node classification, we use \textbf{Logistic Regression (LR)} on raw node features.

\item[Graph-based.] These methods primarily use graph topology. For link prediction, we use the \textbf{Jaccard Coefficient}, which scores links based on neighbor overlap: $\frac{|\mathcal{N}(u) \cap \mathcal{N}(v)|}{|\mathcal{N}(u) \cup \mathcal{N}(v)|}$. For node classification, we use \textbf{Label Propagation (LP)}, a semi-supervised algorithm that diffuses labels from known to unknown nodes. For subgraph regression, we use \textbf{WL-Hash} \cite{shervashidze2011weisfeiler}, a feature vector derived from the Weisfeiler-Lehman test that captures local neighborhood structure.

\item[GNN-based.] This serves as a practical \textbf{supervised reference point} for node- and link-level tasks. We use a \textbf{Supervised GraphSAGE} model trained end-to-end with full access to the task labels, optimized with early stopping on the validation set. We omit a GNN baseline for subgraph regression, as this would require a distinct graph-level regression architecture (e.g., a batched GNN with graph pooling) that is not directly comparable to our simple linear probe methodology.
\end{description}

We report test metrics and compute paired bootstrap significance tests ($n=2000$, $\alpha=0.05$) comparing "Ours" versus the best baseline. All baselines use identical data splits; augmentations apply only during SSL pretraining.

For each test fold, we form positives from the held-out $15\%$ edges and sample an equal number of negatives uniformly over non-edges, rejecting self-loops and duplicates until the target count is reached. Negatives are drawn disjoint from all observed positives, and the train/val negatives used by the logistic probe are disjoint from the test pairs.

\paragraph{Stage 3: Transfer learning.}
We generate 500 graphs with varying $N,K$ and construct hand-crafted graph-level features $X_{\text{graph}}$ (means/stds of node and edge features plus simple structural summaries). We classify graphs by $K$ using logistic regression (80/20 split) and compare against a baseline using aggregated raw node features.

\paragraph{Stage 4: Ablations.}
We train 9 variants (400 epochs each): FULL, NO\_EDGESET ($\lambda_E=0$), NO\_VARFLOOR, NO\_COV, NO\_EDGE\_HEAD (no edge embeddings), NO\_PREDICTORS, NO\_FEAT\_MASK, NO\_DROPEDGE, NO\_EDGE\_NORM. Each is evaluated on Stage 2 tasks.

\section{Results}

\subsection{Hyperparameter Selection}

\begin{figure}[t]
\centering
\includegraphics[width=0.48\textwidth]{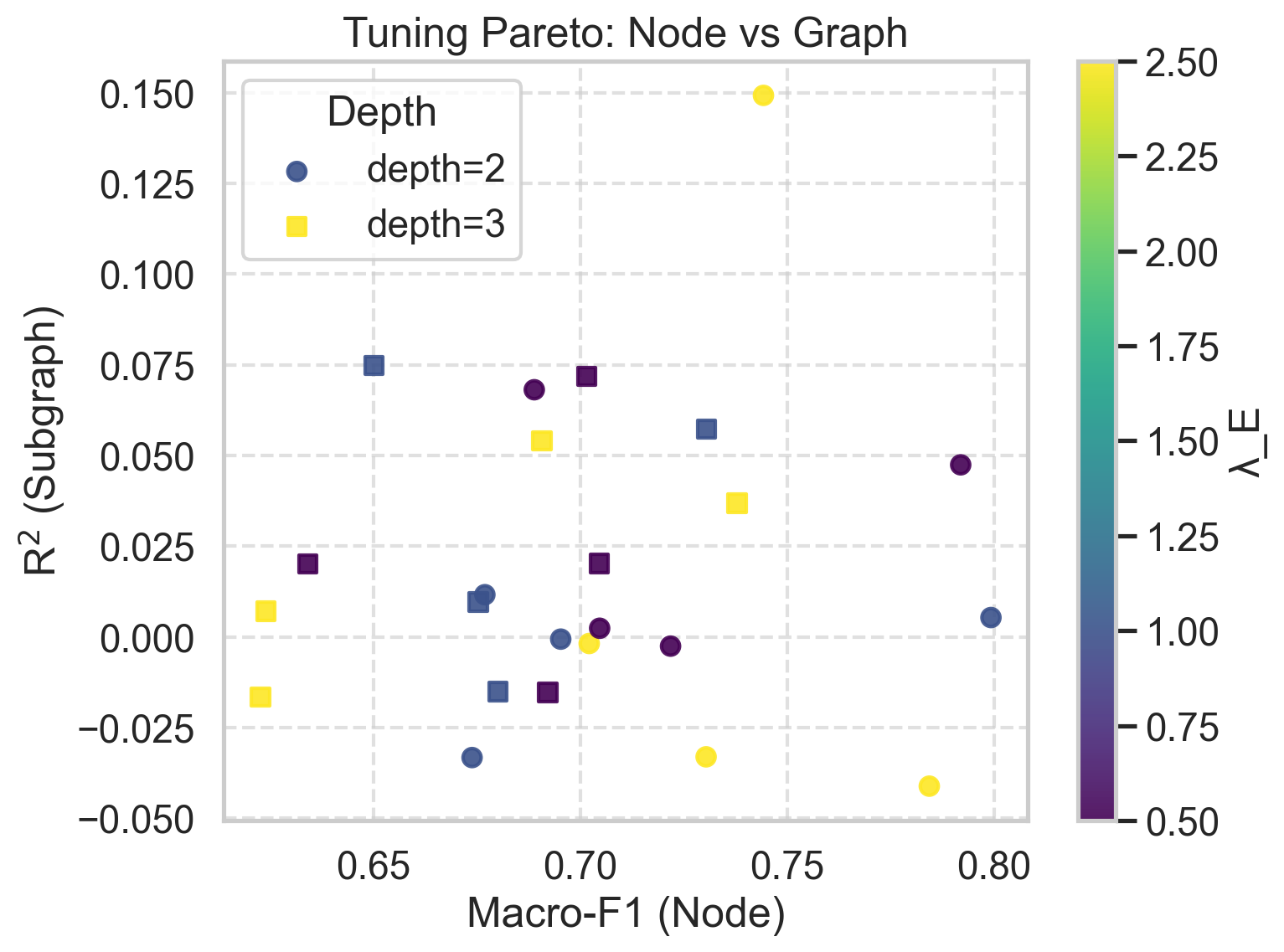}
\caption{\textbf{Hyperparameter trade-offs.} Pareto frontier between node F1 and graph $R^2$ reveals weak trade-off.}
\label{fig:tuning}
\end{figure}

Among 24 configurations, the combination $\text{hidden}=64$, $\text{depth}=2$, $\text{emb\_dim}=64$, and $\lambda_E=2.5$ achieved the highest composite validation score ($S_{\text{comp}}=2.19$). Figure~\ref{fig:tuning} visualizes the trade-off between node-level and graph-level performance. Node classification (x-axis) and subgraph regression (y-axis) show a non-linear relationship: increasing $\lambda_E$ (lighter colors) improves $R^2_{\text{graph}}$ but can slightly reduce node-level macro-F1. This confirms that stronger edge-distribution alignment emphasizes graph-scale consistency at the expense of fine-grained node separability. Both 2-layer (circles) and 3-layer (squares) variants display similar trends, though deeper networks show greater spread in $R^2_{\text{graph}}$, indicating higher representational flexibility but also greater sensitivity to $\lambda_E$.

\begin{figure}[htbp]
\centering
\includegraphics[width=0.48\textwidth]{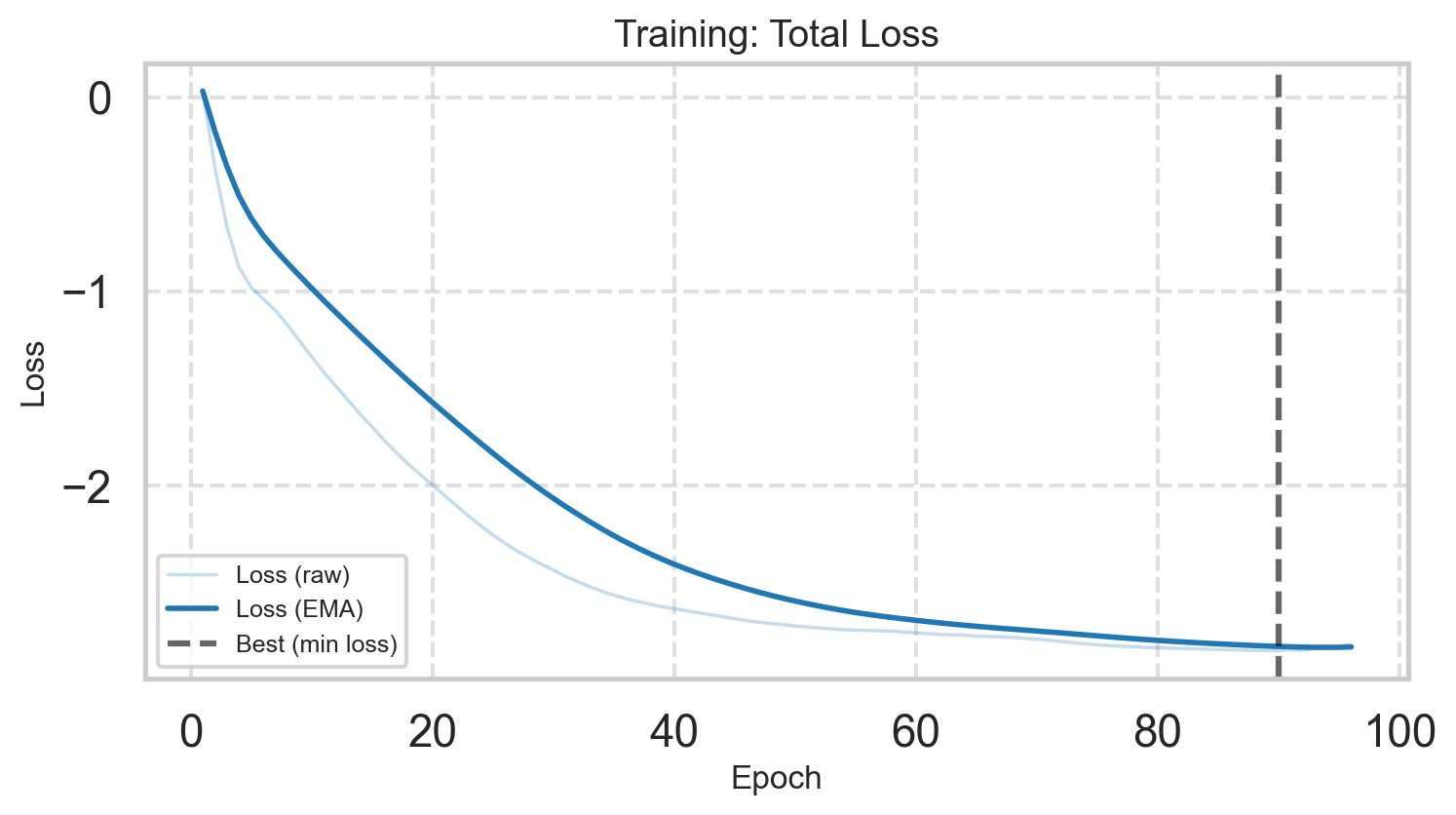}\label{fig:train_loss}
\includegraphics[width=0.48\textwidth]{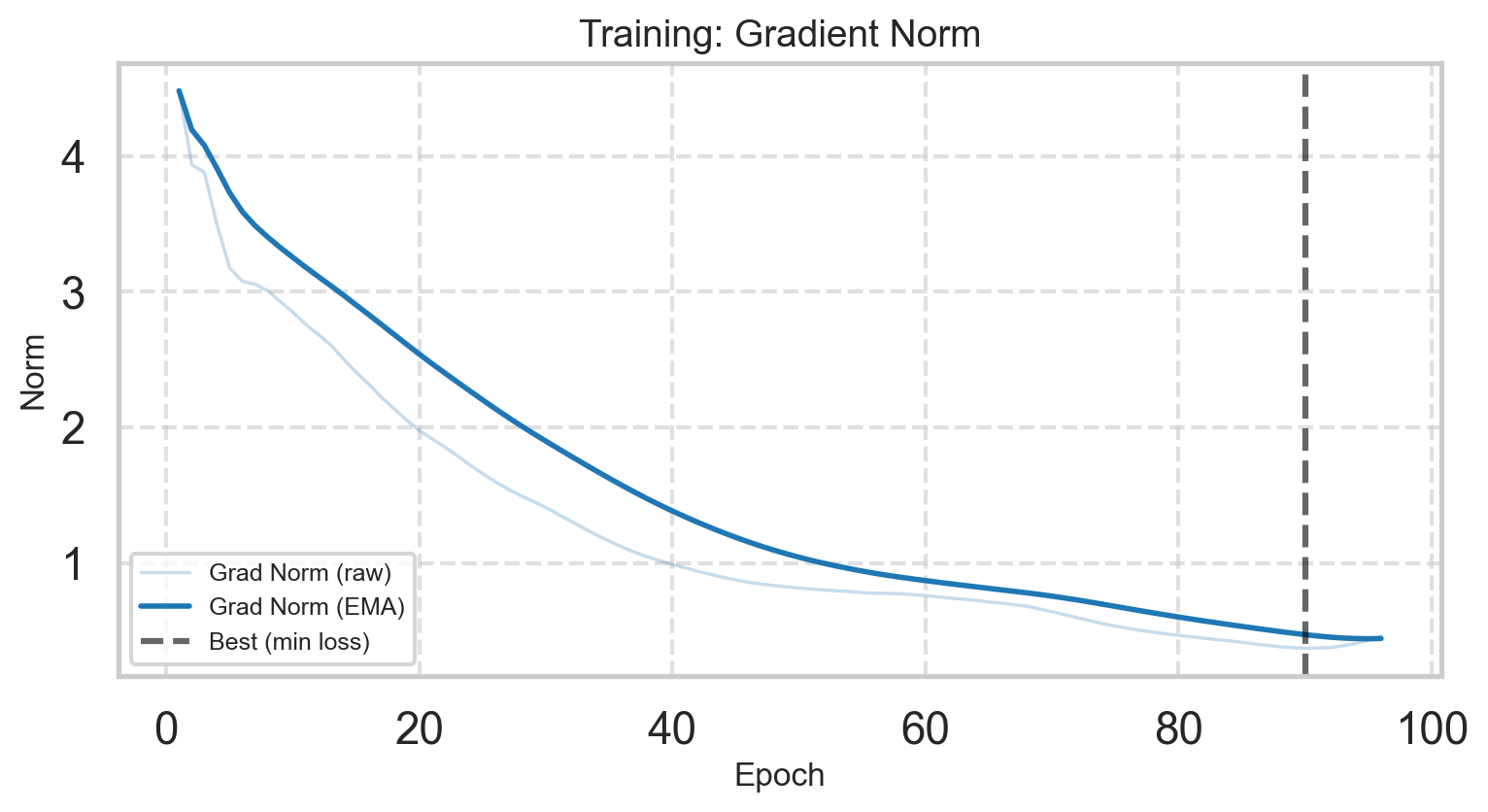}\label{fig:train_grad}
\includegraphics[width=0.48\textwidth]{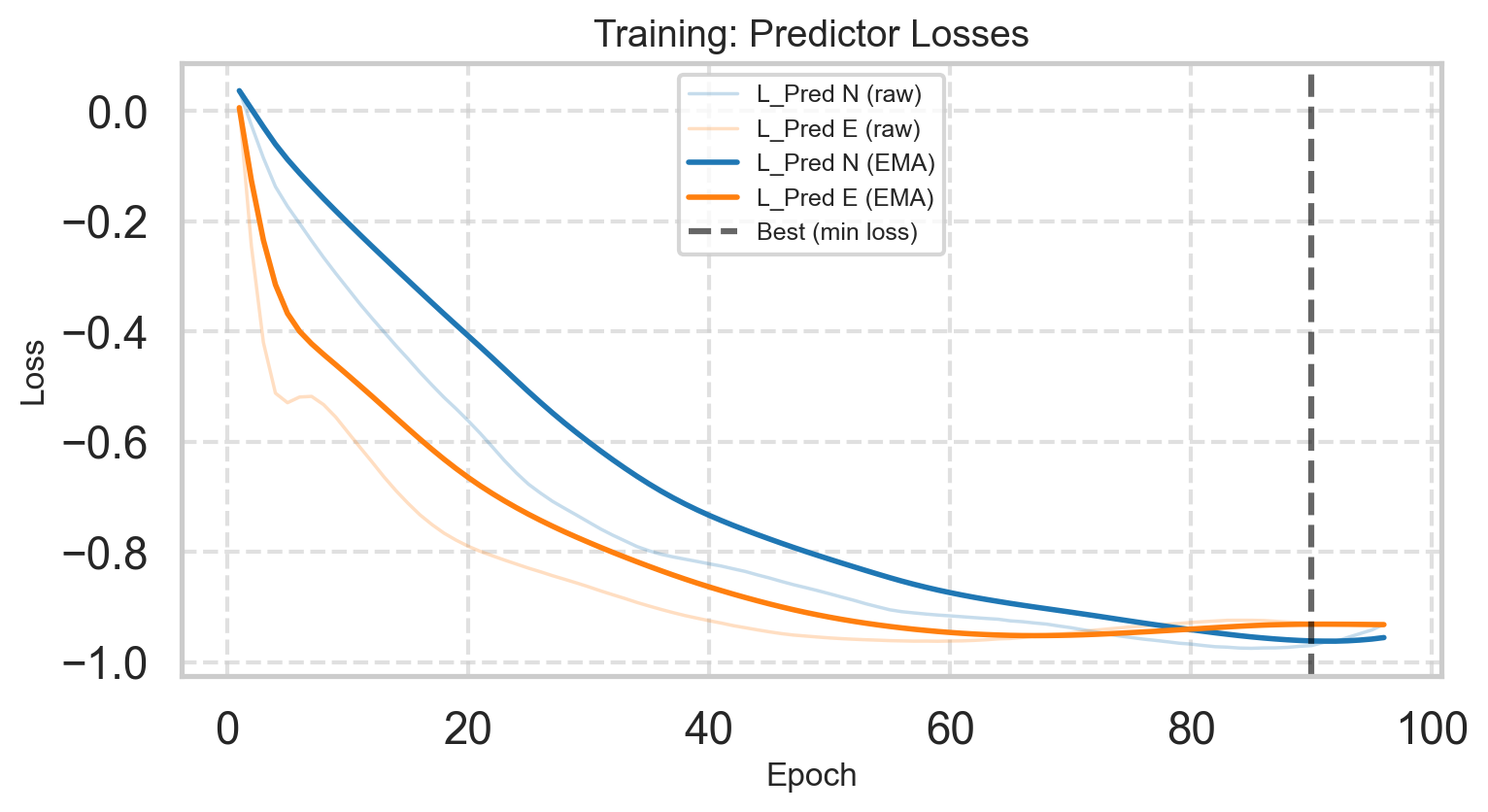}\label{fig:train_pred}
\includegraphics[width=0.48\textwidth]{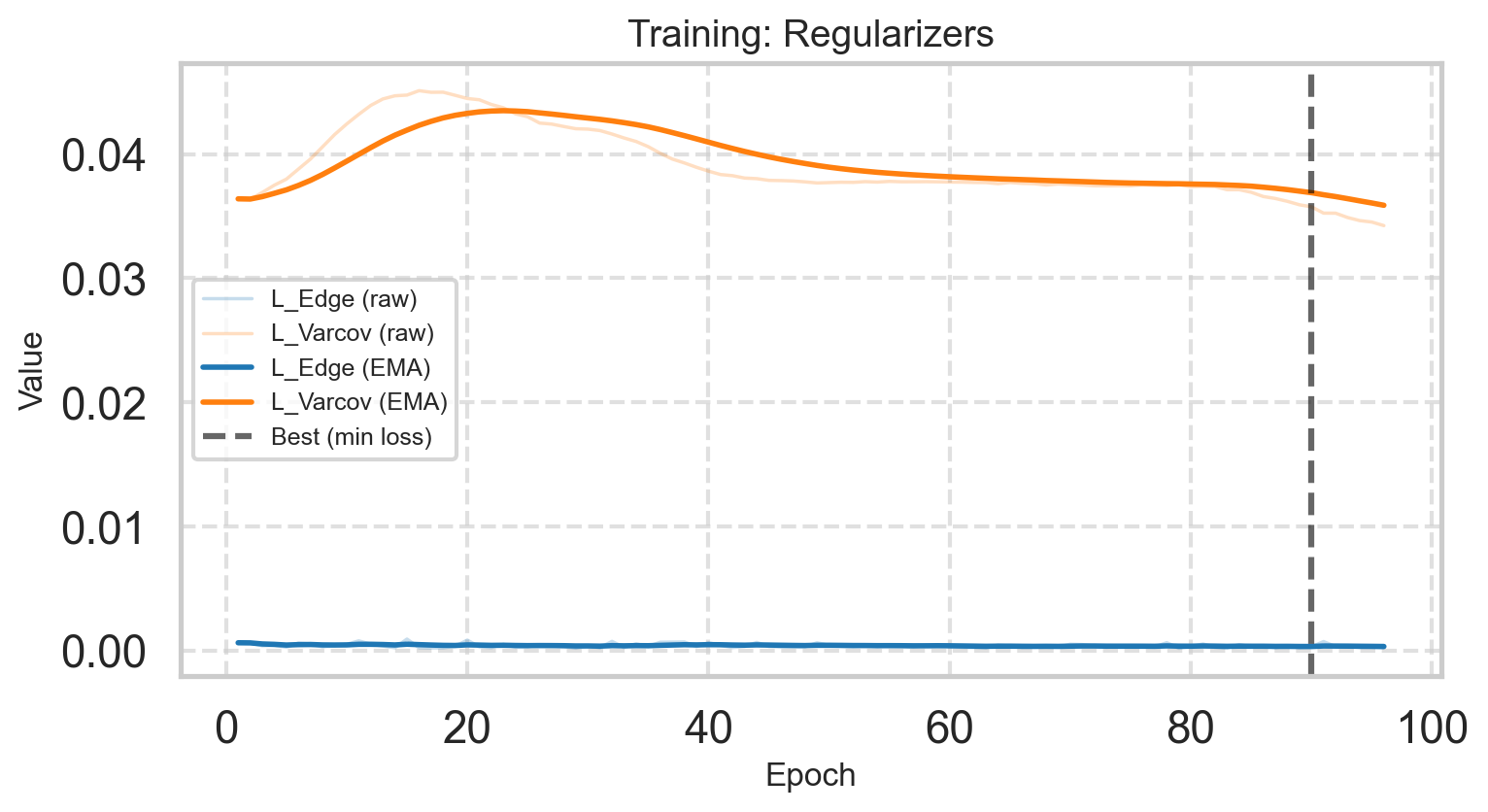}\label{fig:train_reg}
\caption{\textbf{Training Dynamics.} Smoothing for all graphs was done with an exponential moving average (EMA).}
\label{fig:training_loss}
\end{figure}

Figure~\ref{fig:training_loss} summarizes optimization dynamics: (a) total loss decreases steadily until early stopping (best validation loss at $\sim$epoch 90); (b) predictor losses become increasingly negative, indicating improved alignment; (c) MMD stabilizes quickly while VICReg regularization remains active; (d) gradient norms decay smoothly without instability.

\begin{figure}[htbp]
\centering
\includegraphics[width=0.45\textwidth]{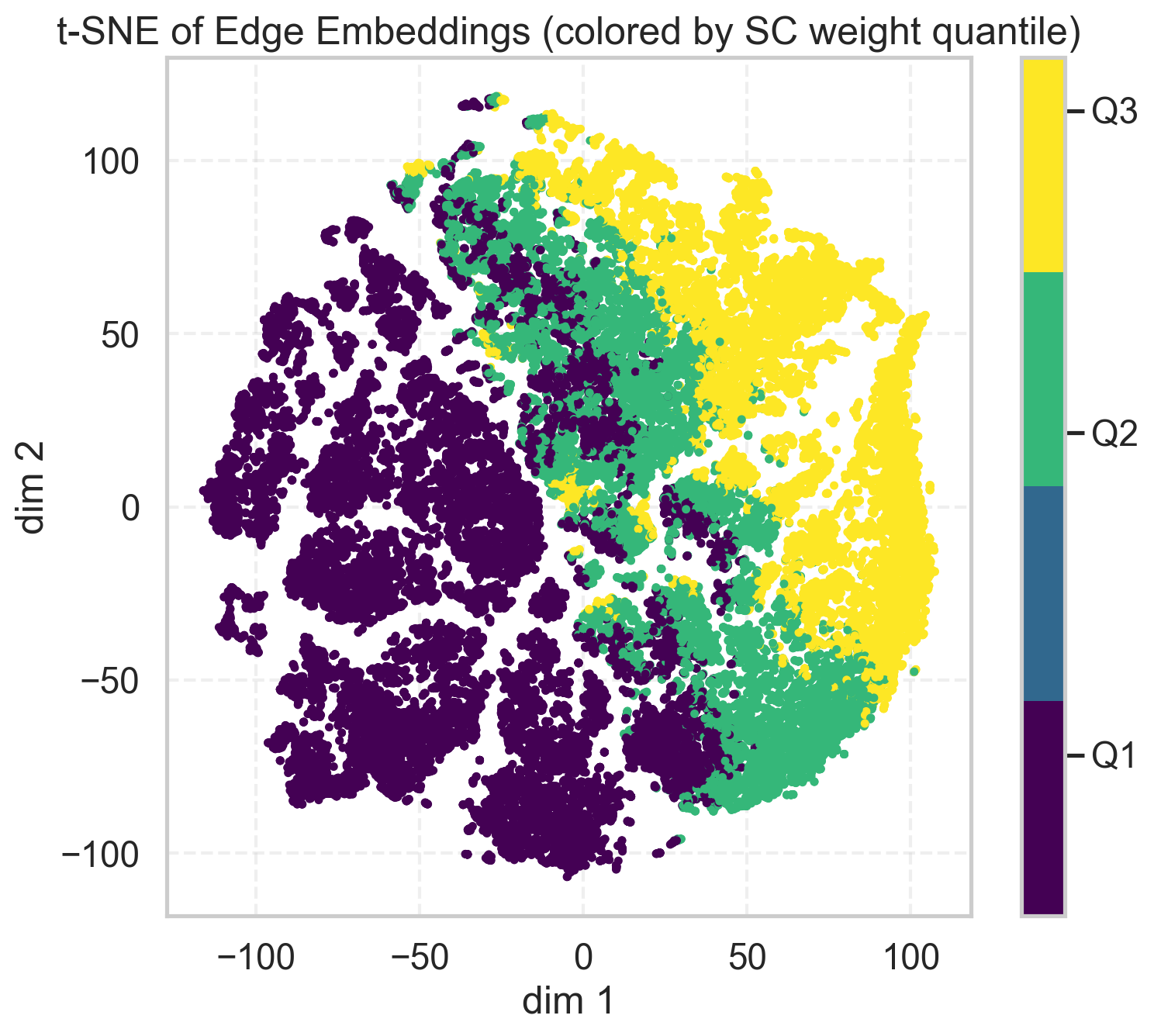}\label{fig:tsne_edges}
\includegraphics[width=0.45\textwidth]{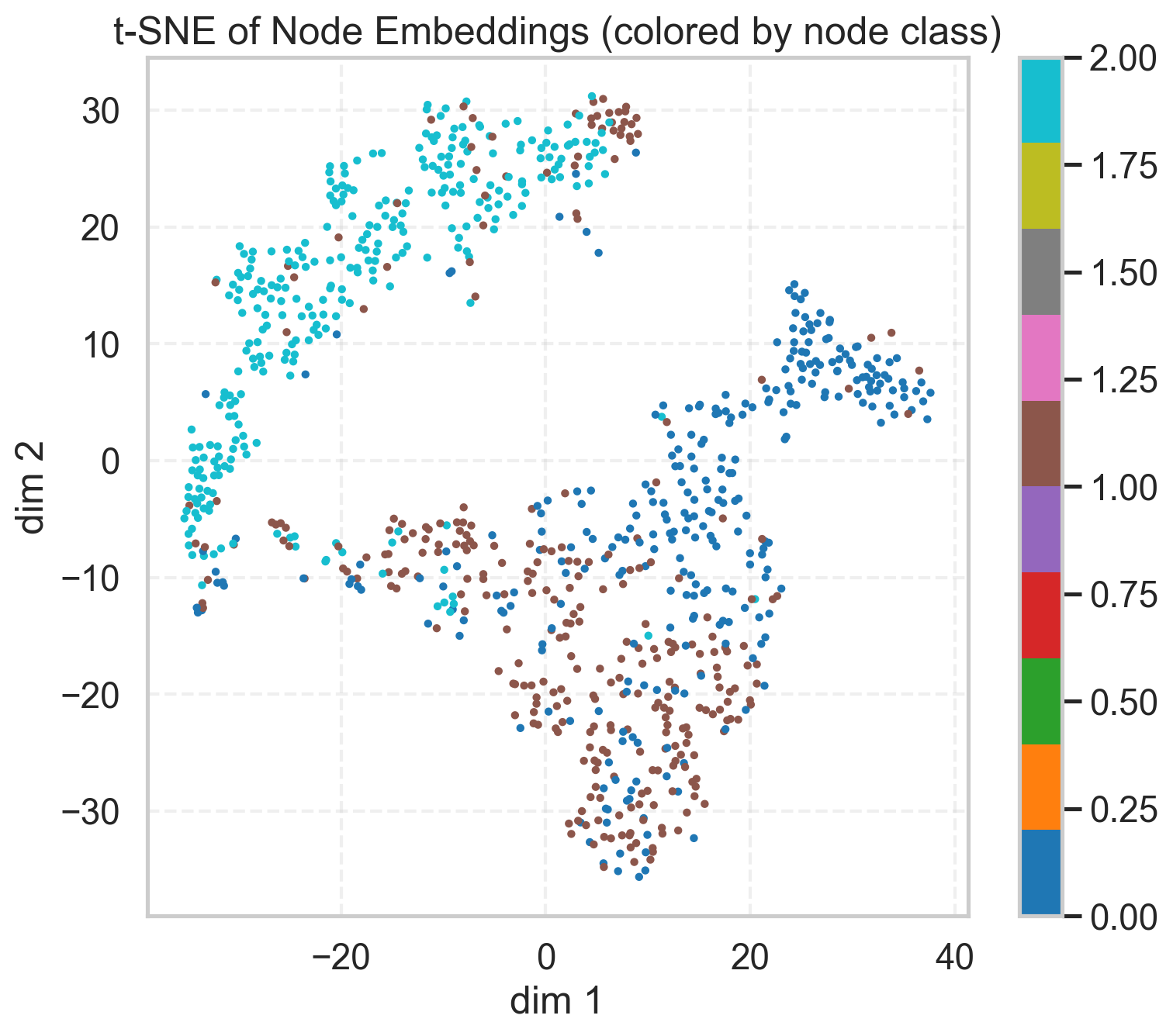}\label{fig:tsne_nodes}
\caption{\textbf{t-SNE of embeddings.} Edges colored by SC weight quartiles show clear clustering. Nodes colored by class show partial separation.}
\label{fig:tsne}
\end{figure}

\begin{description}
\item[Total loss.] The objective decreases steadily and plateaus, with the best validation loss found at $\sim$epoch 90, indicating stable descent without late-stage oscillation.
\item[Predictor losses.] Both SimSiam predictor terms (node and edge) become increasingly negative and flatten over time, consistent with improving cross-view alignment (more negative $\Rightarrow$ better alignment for the cosine-style SimSiam objective).
\item[Regularizers.] The edge-distribution MMD remains small after an initial transient, suggesting the two augmented views quickly yield similarly distributed edge embeddings. The VICReg variance/covariance penalty rises early, peaks, and then slowly decays while staying strictly positive, i.e., it is active and pushes per-dimension variance toward the target while suppressing off-diagonal covariance to avoid collapse.
\item[Gradient norm.] The global gradient norm decays smoothly across epochs, corroborating a well-conditioned optimization without signs of exploding or vanishing gradients.
\end{description}

t-SNE visualizations (Figure~\ref{fig:tsne}) reveal well-organized embedding spaces. Edge embeddings cluster smoothly by structural connectivity (SC) weight quantiles, confirming that the model captures continuous relational properties rather than collapsing to discrete groups. Node embeddings exhibit partial but coherent class separation, consistent with moderate node-classification performance. Together, these visualizations support that the learned representations preserve fine-grained structure while maintaining global smoothness across embedding scales.

\subsection{Single-Graph Downstream Tasks}

Table~\ref{tab:main} summarizes probe performance using frozen embeddings from self-supervised pretraining. Our SSL-pretrained embeddings, when frozen, do not outperform simpler, classical baselines on their corresponding tasks.

\begin{table}[htbp]
\centering
\caption{\textbf{Single-graph probe results.} Symbols indicate significantly lower performance compared to the best-performing baseline for each task: $\dagger$ $p{<}0.001$ and $*$ $p{<}0.01$.}
\label{tab:main}
\small
\begin{tabular}{llcc}
\hline
\textbf{Task} & \textbf{Method} & \textbf{Metric} & \textbf{Score} \\
\hline
Link Pred & Classical: Cosine & AUC & 0.686 \\
Link Pred & Graph: Jaccard & AUC & \textbf{0.845} \\
Link Pred & GNN: SAGE (sup.) & AUC & 0.802 \\
Link Pred & \textbf{Ours}: LR($z_e$) & \textbf{AUC} & \textbf{0.727}$^{\dagger}$ \\
\hline
Node Cls & Classical: LR & F1 & 0.794 \\
Node Cls & Graph: LabelProp & F1 & 0.690 \\
Node Cls & GNN: SAGE (sup.) & F1 & \textbf{0.848} \\
Node Cls & \textbf{Ours}: MLP($z_n$) & \textbf{F1} & \textbf{0.767}$^{*}$ \\
\hline
Subgr Reg & Classical: Ridge(pool) & $R^2$ & \textbf{0.189} \\
Subgr Reg & Graph: WL-Hash & $R^2$ & 0.177 \\
Subgr Reg & \textbf{Ours}: Ridge($z_e$) & \textbf{$R^2$} & \textbf{-0.174}$^{*}$ \\
\hline
\end{tabular}
\end{table}

\paragraph{Link prediction.}
Our model ($AUC{=}0.727$) is catastrophically outperformed by the classical Jaccard coefficient ($AUC{=}0.845$, $p{<}0.001$, $95\%CI=[-0.123,-0.112]$). The Jaccard coefficient succeeds because it is a pure, explicit measure of topological community structure (shared neighbors). Our SSL model, trained to be invariant to augmentations like \texttt{DropEdge}, learns to ignore the precise topological information that Jaccard exploits. This reveals a fundamental objective mismatch between generic SSL and topology-driven connectome analysis.

\paragraph{Node classification.}
Our frozen node embeddings ($F1{=}0.767$) are slightly outperformed by Logistic Regression on raw features ($F1{=}0.794$). Performance remains far below the supervised GraphSAGE upper bound ($F1{=}0.840$, $p{<}0.01$, $95\%CI=[-0.145,-0.023]$). This result further suggests that learned representations offer no clear advantage over the raw features, even for a simple node-level task.

\paragraph{Subgraph regression.}
For subgraph-level prediction, the method fails completely, achieving a negative $R^2$ ($-0.174$) that is significantly worse than the classical Ridge baseline ($R^2{=}0.189$, $p{<}0.01$, $95\%CI=[-0.686,-0.125]$). This indicates that the model's representations are not capturing meaningful structural properties at the subgraph level, further confirming the model's failure to learn topology.

Collectively, these results show that the invariances learned by our hierarchical framework are not well-aligned with the properties required by these downstream tasks, a finding we explore in the ablation study and discussion.

\subsection{Transfer Learning Across Graphs}

\begin{figure}[htbp]
\centering
\includegraphics[width=0.48\textwidth]{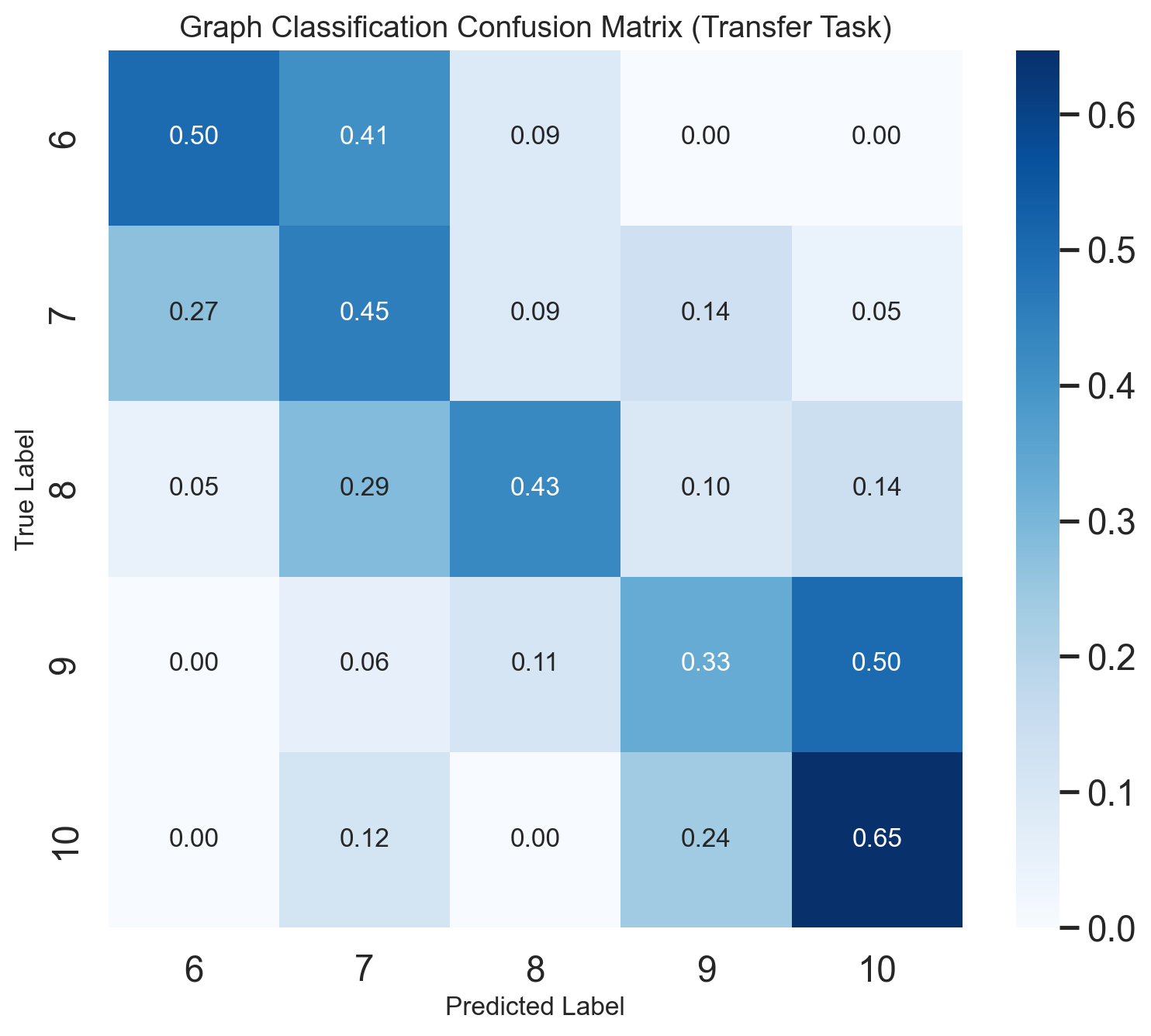}
\caption{\textbf{Transfer task confusion matrix.} Frozen $z_G$ classifies graphs by $K$ (5 classes).}
\label{fig:transfer}
\end{figure}

On the graph classification transfer task, our frozen embeddings achieve $47.2\%$ accuracy, exceeding chance ($20\%$) but underperforming the classical feature-based baseline ($53\%$). The confusion matrix (Figure~\ref{fig:transfer}) shows strong diagonal structure but confusion between adjacent community counts (e.g., $K{=}6$ and $K{=}7$). This indicates that the embeddings preserve coarse modular organization while blurring fine structural differences. Such smooth generalization mirrors cortical representations that encode continuous gradients of organization—capturing topology rather than discrete category boundaries.

\subsection{Ablation Study}

\begin{figure}[htbp]
\centering
\includegraphics[width=0.48\textwidth]{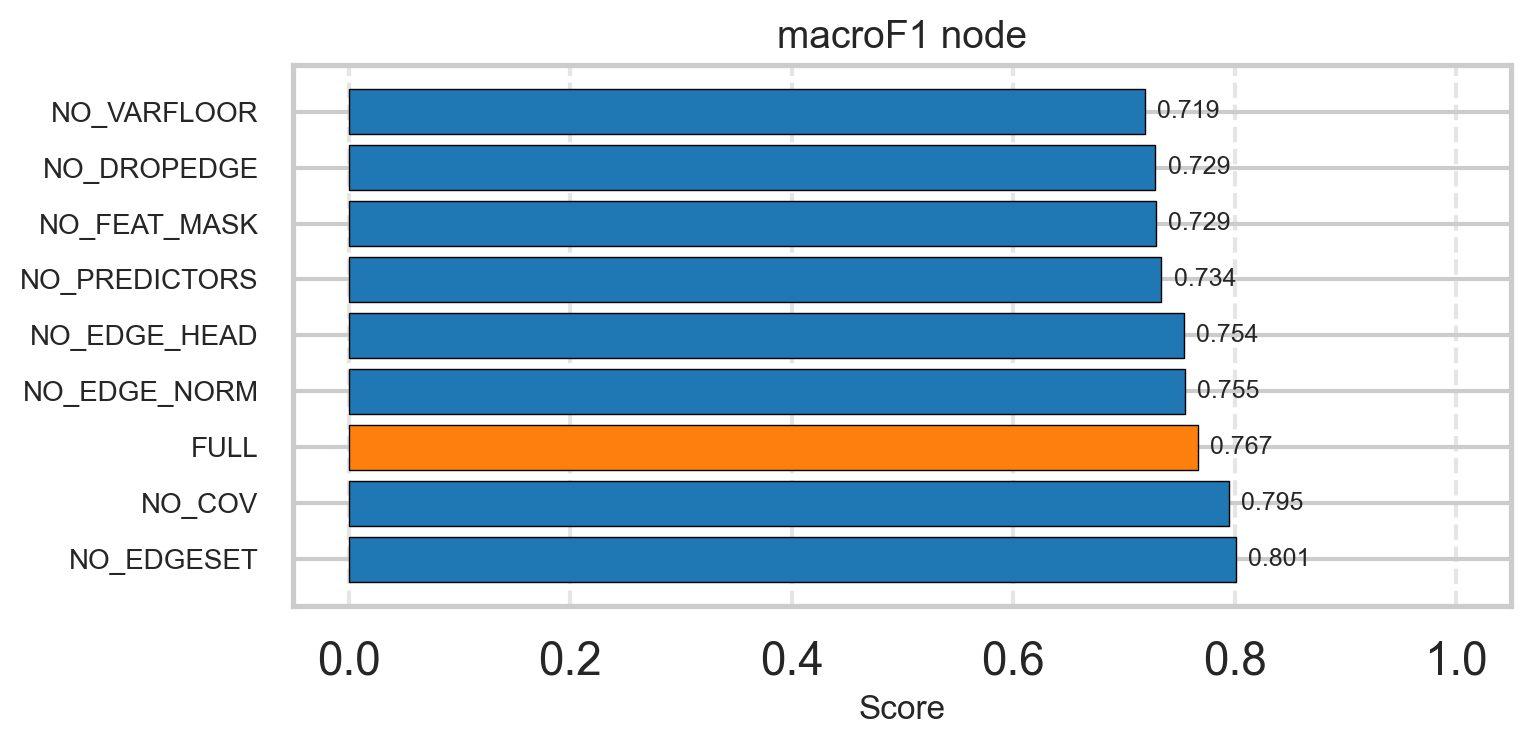}\label{fig:ablation_node}
\includegraphics[width=0.48\textwidth]{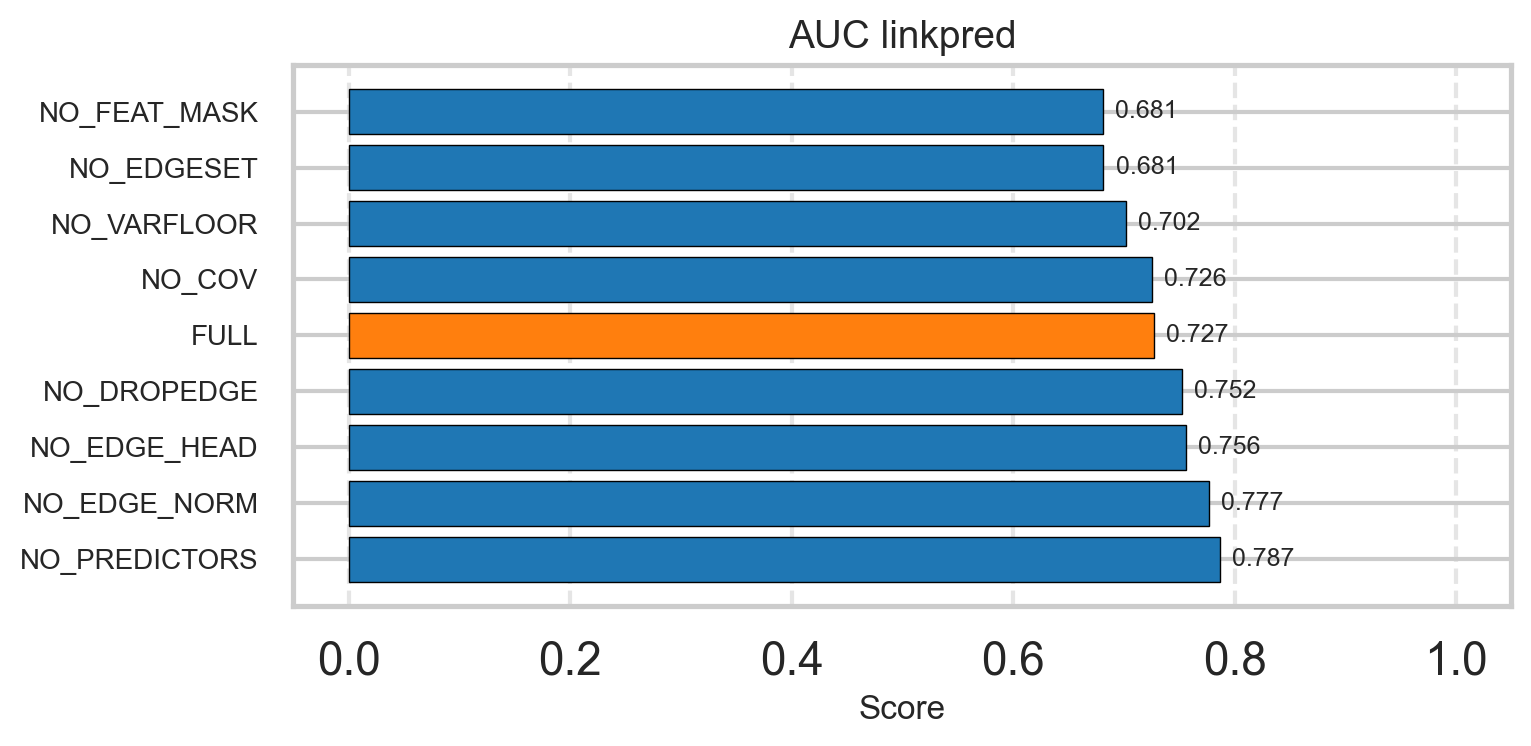}\label{fig:ablation_link}
\includegraphics[width=0.48\textwidth]{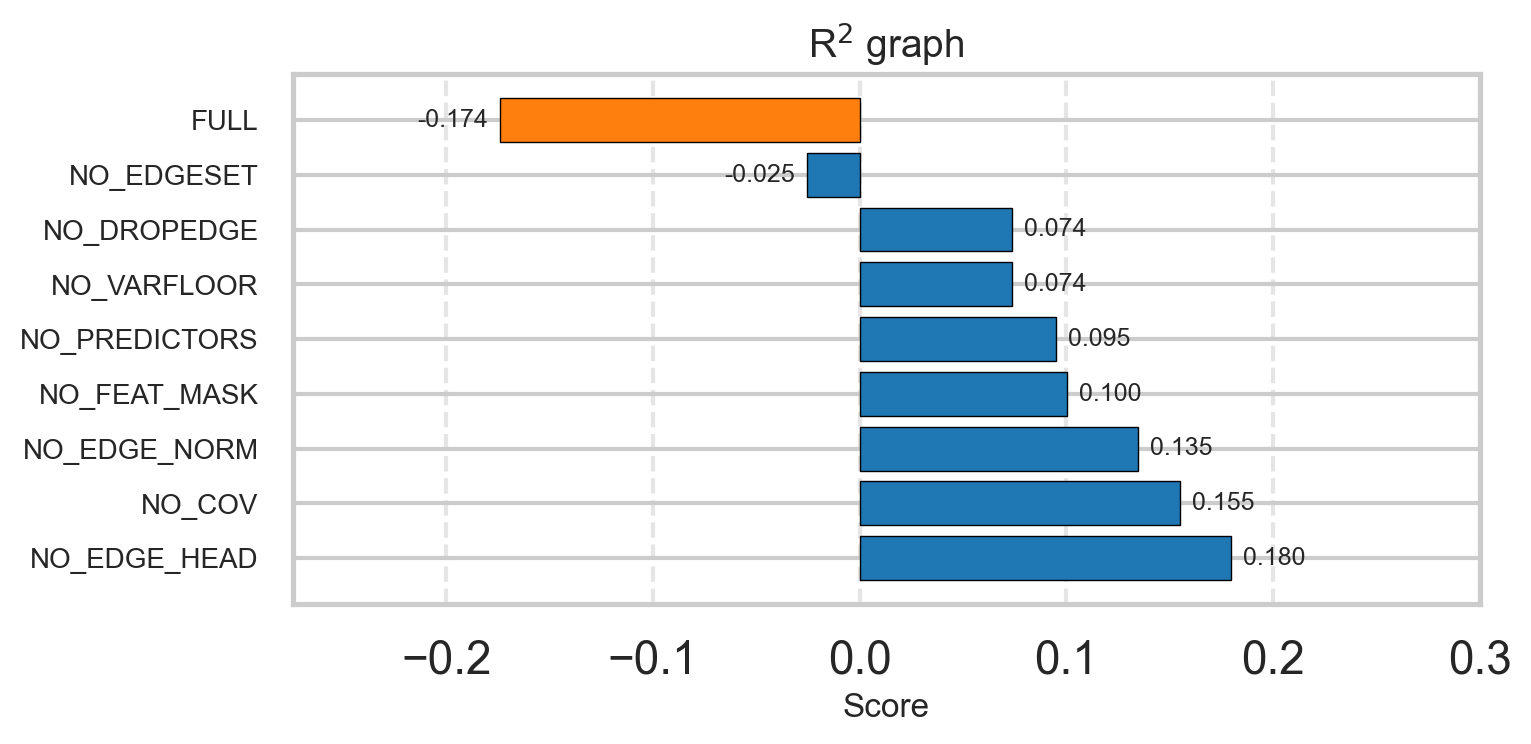}\label{fig:ablation_graph}
\caption{\textbf{Ablation study.} Absolute performance for each metric across variants. FULL is highlighted for reference.}
\label{fig:ablation}
\end{figure}

Figure~\ref{fig:ablation} reveals critical trade-offs, showing that no single configuration excels across tasks.
\begin{itemize}
    \item For \textbf{node classification}, the \texttt{FULL} model ($F1{=}0.767$) is outperformed by variants removing the edge distribution loss (\texttt{NO\_EDGESET}, $F1=0.801$) or covariance penalty (\texttt{NO\_COV}, $F1=0.795$), suggesting the model is over-regularized.    
    \item For \textbf{link prediction}, the \texttt{FULL} model ($AUC{=}0.727$) is one of the worst. Crucially, removing the topological augmentation (\texttt{NO\_DROPEDGE}, $AUC=0.752$) improves performance, directly supporting our hypothesis that this invariance objective is detrimental. The best performance comes from removing the predictors (\texttt{NO\_PREDICTORS}, $AUC=0.787$), pointing to objective misalignment.
    \item For \textbf{subgraph regression}, the \texttt{FULL} model fails completely ($R^2{=}-0.174$). Here too, \texttt{NO\_DROPEDGE} ($R^2=0.074$) provides a substantial improvement, reinforcing the harm of the topological invariance. The best score comes from removing the architectural \texttt{NO\_EDGE\_HEAD} ($R^2=0.180$), indicating our multi-task design is an unstable compromise.
\end{itemize}

Together, these results strongly support our failure analysis. The \texttt{FULL} model is an unstable compromise between competing, misaligned objectives. The results show that no amount of tuning these specific components can fix the fundamental mismatch between generic, invariance-based SSL and the topology-driven tasks essential for graph analysis in neuroscience.

\section{Discussion}

Our hierarchical SSL framework was designed to apply multi-scale SSL to connectome-like graphs. The model's consistent failure to outperform simple heuristics is the most valuable finding of this study. The evaluation reveals a fundamental mismatch between generic SSL objectives and the properties of neuro-inspired graphs.

\paragraph{Failure Analysis: Invariance Objectives vs. Topological Structure.}
Our most critical result is the model's failure against the Jaccard coefficient, indicating an objective mismatch. The Jaccard heuristic is a pure, explicit measure of community structure (i.e., shared neighbors). In contrast, modern SSL methods, including ours, are dominated by invariance objectives. By training the model to be invariant to augmentations like \texttt{DropEdge} (which alters the graph's topology), we are effectively teaching the model to ignore the precise structural patterns that Jaccard exploits. Our results serve as a cautionary tale: applying generic, feature-centric SSL to connectome-like graphs is likely to fail because the essential properties of these graphs are topological, not feature-based.

\paragraph{Ablation Analysis: A Harmful Objective and Unstable Architecture.}
The ablation study provides direct support for this objective-mismatch hypothesis. The most critical finding is from the \texttt{NO\_DROPEDGE} variant: removing this topological augmentation---the very component that teaches invariance---improves performance on both link prediction ($AUC=0.752$ vs. $0.727$) and subgraph regression ($R^2=0.074$ vs. $-0.174$). This confirms that the invariance objective is not just neutral but actively detrimental to learning topological properties. The ablations also reveal a secondary issue: the \texttt{FULL} model is an unstable compromise of competing components. The fact that other variants, like \texttt{NO\_EDGE\_HEAD} (for subgraph regression) or \texttt{NO\_PREDICTORS} (for link prediction), achieve the best scores on specific tasks demonstrates that the architectural components are misaligned.

\paragraph{Synthetic Evaluation as Controlled Neuroscience Simulation.}
The synthetic benchmark was deliberately designed to emulate core properties of connectomic data: multimodal node features, multi-channel edges, and modular community structure. Although synthetic evaluation limits ecological validity, it parallels the use of model organisms in neuroscience: simplified systems that enable precise hypothesis testing free from the irreducible confounds of biological noise and acquisition artifacts. Future validation on real connectomes (e.g., HCP, UK Biobank) will be essential. Furthermore, real connectomes exhibit more complex multi-scale organization (e.g., overlapping modules, rich clubs). The clear failure of invariance-based SSL even on our simplified benchmark suggests the objective mismatch problem may be even more pronounced on real, complex brain graphs.

\paragraph{Implications for Neuro-Inspired AI.}
From the \textit{Neuro $\to$ AI} perspective, our failure provides a clear directive: brain-inspired AI must develop objectives that go beyond feature-based invariance. Future models must explicitly reward the preservation of topological properties, such as community structure and small-worldness, rather than treating them as noise to be ignored.

\paragraph{Implications for AI-Driven Neuroscience.}
From the \textit{AI $\to$ Neuro} side, our work is a strong caution against applying ``off-the-shelf'' invariance-based graph self-supervised learning models to connectome data. We show that a similar model may fail to capture the most salient topological properties of brain networks. Our results demonstrate that classical, simpler graph metrics like the Jaccard coefficient remain highly effective and potentially more reliable for certain topology-driven tasks, like link prediction within community structures.

\paragraph{Limitations and Future Directions.}
Several limitations are present. First, all results are derived from synthetic graphs. While this allowed us to isolate the objective mismatch, validation on real connectomes is needed. Second, comparisons excluded recent SSL baselines (e.g., GraphCL, BGRL, GraphMAE). Our critique, however, is not of a specific model but of the invariance-to-topology objective (e.g., \texttt{DropEdge}) that is central to this entire paradigm. Our finding that this objective fundamentally conflicts with topology-based tasks (where Jaccard excels) strongly suggests these methods would exhibit similar failures. Therefore, the critical future direction is developing new, topology-aware pre-training objectives. This could include pretext tasks like graph motif or community prediction, or objectives that explicitly reward the preservation of graph-theoretic properties (e.g., modularity, clustering coefficients) across augmentations.

\section{Conclusion}

We presented a hierarchical SSL framework inspired by multimodal brain networks. Our evaluation on a controlled, connectome-like synthetic benchmark revealed a critical failure: the model was consistently outperformed by simple, classical heuristics. We traced this failure to a fundamental objective mismatch where modern invariance-based SSL trains models to ignore the rich topological and community structure that is the hallmark of brain-like graphs.

Rather than an algorithmic shortcoming, we present this failure as a cautionary and constructive finding. Our synthetic benchmark, validated as a ``model organism'' exhibiting small-world properties, serves as a testbed that exposes the limitations of current SSL. Our analysis indicates that a path to robust graph models in neuroscience requires moving beyond generic invariance.

Looking forward, this work highlights the need for new, topology-aware pre-training objectives. For \textit{Neuro $\to$ AI}, models must be designed to explicitly reward the preservation of structural organization (e.g., modularity or motifs), not discard it as noise. For \textit{AI $\to$ Neuro}, researchers must be cautious of applying off-the-shelf SSL models, as classical, topology-based metrics remain more robust for analyzing connectome structure. By highlighting this critical pitfall, we hope to guide future research toward developing graph models that are truly "brain-inspired" in their objectives.

\bibliography{aaai2026}


\end{document}